%% file: anonymous-submission-latex-2024.tex
\documentclass[letterpaper]{article} 
\usepackage{aaai24}  
\nocopyright 
\usepackage{times}  
\usepackage{helvet}  
\usepackage{courier}  
\usepackage[hyphens]{url}  
\usepackage{graphicx} 
\usepackage{natbib}  
\usepackage{caption} 
\usepackage[ruled]{algorithm2e}

\usepackage{makecell}
\usepackage{booktabs}
\usepackage{multirow}
\usepackage{multicol}
\usepackage{xcolor}
\usepackage{amsmath,amsfonts}
\newcommand{\ours}{\mbox{SGL}\space}
\newcommand{\ourpts}{\mbox{SGL-PT}\space}
\newcommand{\our}{\mbox{SGL}}
\newcommand{\ourpt}{\mbox{SGL-PT}}
\newcommand{\eg}{\textit{e.g.}}
\newcommand{\ie}{\textit{i.e.}}

\usepackage{newfloat}
\usepackage{listings}
\DeclareCaptionStyle{ruled}{labelfont=normalfont,labelsep=colon,strut=off} 
\title{SGL-PT: A Strong Graph Learner with Graph Prompt Tuning}
\author{
    Yun Zhu,
    Jianhao Guo,
    Siliang Tang,
}
\affiliations{
    Zhejiang University\\

    \{zhuyun\_dcd, guojianhao, siliang\}@zju.edu.cn
}

\usepackage{bibentry}

\begin{document}
\maketitle

\input{sections/abstract}
\input{sections/intro}
\input{sections/related}
\input{sections/method}
\input{sections/experiments}

\input{sections/conclusion}
\newpage
\bibliography{aaai24}
\newpage
\input{sections/appendix}

\end{document}

%% file: sections/abstract.tex
\begin{abstract}
Recently, much exertion has been paid to design graph self-supervised methods to obtain generalized pre-trained models, and adapt pre-trained models onto downstream tasks through standard fine-tuning. But the gap between pretext and downstream tasks can limit pre-trained models' potential, leading to negative transfer. Meanwhile, prompt tuning has seen emerging success in natural language processing (NLP) by aligning pre-training and fine-tuning with consistent training objectives. In this paper, we identify the challenges for graph prompt tuning:  
The first is the lack of a strong and universal pre-training task across sundry pre-training methods in graph domain. Such a task should be easily emulated by downstream tasks, akin to the Masked Language Modeling (MLM) in the NLP domain. 
The second challenge lies in the difficulty of designing a consistent training objective for both pre-training and downstream tasks due to the inherent abstraction of graph data. To overcome above obstacles, we propose a novel framework named \ourpts which follows the learning strategy ``Pre-train, Prompt, and Predict''. Specifically, we raise a strong and universal pre-training task coined as \ours that acquires the complementary merits of generative and contrastive self-supervised graph learning. 
And motivated by prompt design in NLP, we reformulate the downstream task as maksed node prediction by designing a novel verbalizer-free prompting function, resulting in unifying the objectives of pre-text and downstream tasks.
Empirical results show that our pre-training method surpasses other baselines under unsupervised setting, and our prompt tuning method can significantly facilitate models on biological datasets over standard fine-tuning and other graph prompt methods.
    
\end{abstract}

%% file: sections/intro.tex
\section{Introduction}
Graph self-supervised Learning methods \cite{gmae,you2020graph} have emerged to create generalized pre-trained models without labels.
To adapt these models for downstream tasks, a ``pre-train, fine-tune'' approach is often used. However, there exists a gap between pre-training and downstream tasks, that hinders knowledge transfer and potentially leads to negative transfer \cite{negative} or overfitting \cite{zhu2021shift}.
For instance, edge prediction \cite{hu2020strategies} pre-trained on local node relations may struggle to generalize to graph classification tasks requiring global relations. This shift in representation can lead to negative transfer \cite{gppt}. 

A feasible solution for mitigating this gap is to unify the pre-training and fine-tuning models with consistent training objectives by prompt technique~\cite{prompt-survey}. 
Though already mature in NLP domain, graph prompt tuning is still under exploration. We identify the main \emph{challenges} of graph prompt tuning: (1) The need for a strong and universal pre-training task: graph prompting method requires the pre-training task to capture rich information (\eg, intra-data and inter-data relations). Existing pre-training methods fail to learn rich information in graph domain, because most of them only focus on learning local relations~\cite{gmae} or global relations~\cite{you2020graph} while neglecting the inter-dependency of both. Furthermore, graph prompt methods mandate a pre-training task that can be readily emulated, ensuring smooth integration into downstream tasks. However, a pre-training task with these desired characteristics is currently absent.
(2) The difficulty of reformulating the downstream task in the same format of the pre-training task: unlike the cloze template in NLP domain, how to design meaningful prompt templates and verbalizers for graphs remains an open problem due to the inherent abstraction of graph data.
Recently, some works made first attempts on graph prompt tuning. GPPT~\cite{gppt} and GraphPrompt~\cite{graphprompt} unify the pretext and downstream task in a similar format (\ie, edge prediction task). However, edge prediction is a trivial binary classification task, which can not capture rich inter-data information and lose its generality (not solving \emph{challenge 1}). GPF~\cite{gpf} and Prompt Graph (ProG)~\cite{sun2023all} introduce a learnable prompt feature and a learnable prompt graph into the input space as plug-ins, which can be incorporated into any pre-trained models. These methods are more likely new transformation tricks added to input space, while they do not holistically align the training objectives between pretext and downstream tasks, hence not fully addressing \emph{challenge 2}. In short, the challenges we propose are waiting to be solved.

To overcome the above obstacles, we propose a novel graph learning framework coined as \ourpts that follows the ``Pre-train, Prompt and Predict'' strategy. 
Specifically, for \emph{challenge 1}, we design a strong and universal graph self-supervised method named Strong Graph Learner (\our). It combines generative and contrastive models for complementary strengths: the generative method has better robustness and can ratiocinate the characters of nodes according to neighbor through learning local (intra-data) relations, but lacks discriminative representations; The contrastive method focuses on learning qualified global (inter-data) representations through instance discrimination but may lose detailed information on individual graphs. 
We effectively combine these two self-supervised methods through asymmetric design and a dynamic queue to obtain a strong and universal self-supervised graph learner. Besides, this pre-training task can be mimicked by the downstream task painlessly.
For \emph{challenge 2}, we design a novel prompting function that introduces a masked super node into individual graphs and reformulates the downstream graph classification as masked node prediction. 
We realize \emph{verbalizer-free} class mapping by introducing supervised graph prototypical contrastive learning to establish the mapping between reconstructed features and semantic labels. In this way, we unify the training objectives between pretext and downstream tasks. Results show that our pre-training method surpasses other strong generative methods~\cite{gmae} and contrastive baselines~\cite{infogcl}. And our prompt tuning strategy can greatly facilitate models on biological datasets compared to standard fine-tuning and other graph prompt methods.

Our contributions are summarized as follows: 
\begin{itemize}
    \item We identify the main challenges for graph prompting. And we propose a novel framework following ``Pre-train, Prompt, and Predict'' strategy to solve the challenges.
    \item We propose a strong and universal graph self-supervised method SGL unifying generative and contrastive merits through the asymmetric design. And we unify pre-training and fine-tuning by designing a novel verbalizer-free prompting function.
    \item Empirical results show that our method surpasses other baselines under the unsupervised setting, and our prompt tuning method can significantly facilitate models on biological datasets compared to fine-tuning methods.
\end{itemize}

%% file: sections/related.tex
\section{Related Works}
\subsection{Graph Self-supervised Learning}
Graph self-supervised methods can be classified into three categories: Predictive, Generative and Contrastive~\cite{wu2021self}. 
Predictive method  self-generates labels by statistical analysis and designs prediction-based pre-training tasks on the generated labels~\cite{jin2020self}. 

Generative method focuses on learning local (intra-data) relations based on pretext tasks such as feature/edge reconstruction~\cite{kipf2016variational,gmae}. 
Recently, the generative-based masked autoencoders~(\eg, GraphMAE~\cite{gmae}) show their superior performance across different tasks. Masked autoencoders learn hidden representations by masking partial node features to obtain high-level representations and then reconstructing these masked node features through decoders.
These methods focus on the local relations between nodes, which may be incompetent for learning discriminative graph representations and deficient for graph classification task.

Contrastive method maximizes the agreement between positive data pairs and pushes away the negative data pairs in the representation space to learn the global~(inter-data) relations between graphs~\cite{you2020graph,rosa}, but may lose the detailed information within a single graph. 

In our work, we combine the generative and contrastive methods through asymmetric design and a dynamic queue for complementary merits to obtain a strong and universal pre-training method.

\subsection{Prompt-based Learning}
The training strategy ``Pre-train and Fine-tune'' is widely used to adapt pre-trained models onto specific downstream tasks. 
However, this strategy ignores the inherent representation gap between pre-training and downstream tasks and leads to poor performance on few-shot problems.

Prompt-based learning~\cite{prompt-survey} is a technique arising from NLP to narrow the gap between pre-training and downstream tasks. It reformulates the downstream task to match the format of the pre-training task. This involves designing a \emph{prompt template} that converts the task into a masked word prediction task. For instance, in sentiment analysis, the label prediction task can be transformed into a masked word prediction by the pre-defined template like ``[X]. It is a [MASK] movie'' for the input sentence [X]=``I love this movie.'', aligning with the pre-training mask language model. The second aspect is the \emph{verbalizer design}, which maps the output word [MASK] to a specific label. For example, words like `good, great' may be associated with the label `+', while `bad, terrible' are associated with the label `-'. 
The dot product of [MASK] for these tokens provides confidence in assigning the sentence a label.

While prompt-based learning is well-established in NLP, it's relatively new in the graph domain. Existing approaches like GPPT~\cite{gppt} and GraphPrompt~\cite{graphprompt} rely on edge prediction as a pre-training task and reformulate the downstream task as edge prediction. But edge prediction is a trivial binary classification task and focuses on learning local (intra-data) relations which will be incompetent for graph classification. GPF~\cite{gpf} introduces a universal prompt feature for various pre-trained models, and ProG~\cite{sun2023all} extends GPF and introduces a universal graph prompt (\ie, a set of prompt features), but they do not unify the training objectives of the pertaining and downstream tasks, limiting its potential.


In contrast to prior works~\cite{gppt,graphprompt}, we address two challenges simultaneously. By considering graph-level task characteristics and the prompt template's dependence on the pretext task, we redesign the pre-training method and graph prompt template to mitigate representation gaps. 
In the following section, we will demonstrate how we overcome these obstacles. 

%% file: sections/method.tex
\section{Method}
\subsection{Preliminaries}
This work focuses on graph classification task on a series of graphs $\mathbb{G}=\left\{G_1, G_2, ..., G_K\right\}$. $G_k=\left(\mathbf{X}_k,\mathbf{A}_k,y_k\right)$ denotes a single graph, where  $\mathbf{X}_k \in \mathbb{R}^{N_k\times D}$ is the raw node features, $\mathbf{A}_k \in \mathbb{R}^{N_k \times N_k}$ is the adjacent matrix, and $y_k$ is the graph label.
For notations, $N_k$ represents node number of graph $k$,  $D$ represents the dimension of raw features, and $\mathbf{A}_k(i,j)=1$ means there exists an edge between node $i$ and node $j$ in graph $k$, otherwise 0.
$\mathcal{E}(\cdot)$ represent GNN encoder and $\mathcal{D}(\cdot)$ represents decoder. $f(\cdot)$ , and $\mathcal{R}(\cdot)$ represent projection head and readout function respectively.
Under unsupervised training setting of this work, the label information $y$ is unavailable for each graph during pre-training.
\begin{figure}[htp]
    \centering
    \includegraphics[scale=0.52]{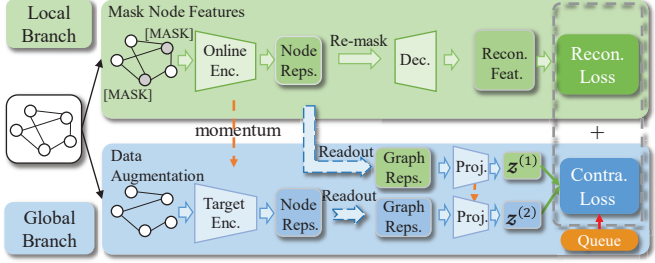}
    \vspace{-1.5em}
    \caption{The pre-training framework:Strong Graph Learner. }
    \label{fig:pretrain}
    \vspace{-1.5em}
\end{figure}

\subsection{SGL: Strong Graph Learner for Pre-training} \label{sec:pretrain}
In this part, we will introduce our proposed pre-training strategy depicted in Figure~\ref{fig:pretrain}, which consists of two branches: \emph{local} and \emph{global}. 
For local branch, it focuses on learning intra-data relations via a graph masked autoencoder. 
For global branch, it empowers the pre-training model with instance-wise discriminative ability by graph contrastive learning.
Then we propose a non-trivial solution to integrate these two branches effectively.
Algorithm~\ref{alg:pretrain} in Appendix A gives more details about the procedure.

\paragraph{Local Branch} In this branch, firstly we mask partial nodes' features and then obtain the high-level node representations $\mathbf{H}^{(1)}$ through online encoder $\mathcal{E}_o$. 
Then, we re-mask the node representations and reconstruct masked node features through the decoder $\mathcal{D}$, this is because GraphMAE~\cite{gmae} empirically found that re-masking the node representations $\mathbf{H}^{(1)}$ for decoding will bring performance improvement. 
Finally, we use scaled cosine error as the criterion. 
Our local branch loss is formulated as:

\begin{equation}
    \mathcal{L}_{\text{local}}=\frac{1}{|\widetilde{\mathcal{V}}|} \sum_{v_i \in \widetilde{\mathcal{V}}}\left(1-\frac{x_i^T \tilde{x}_i}{\left\|x_i\right\| \cdot\left\|\tilde{x}_i\right\|}\right)^\gamma, \gamma \geq 1,
\label{equ:sce}
\end{equation}
where $\widetilde{\mathcal{V}}$ represents masked nodes' set and $\tilde{x}$ denotes reconstructed masked node features. And $\gamma$ is a scaling factor that adjusts the contribution of each sample. This loss is averaged on the masked nodes' set $\widetilde{\mathcal{V}}$. In order to keep the notation uncluttered, we consider the batch size as 1 here.

In this way, local branch has better robustness and can ratiocinate the characters of nodes according to neighbors. However, this method mainly focuses on learning local relations on individual graph, which is incompetent to learn discriminative representations for graph classification. 

\paragraph{Global Branch} To complement the insufficiency of the local branch and make the encoder capture global (inter-data) discriminative information among graphs, we appeal to graph contrastive learning. Even though data augmentation is an essential part of contrastive learning~\cite{you2020graph,raft,rosa}, we empirically found simple augmentations like node feature masking and edge removing are good enough to improve the representation ability of the proposed Strong Graph Learner.

The processes of the global branch can be concluded as: firstly we obtain node representations $\mathbf{H}_k^{(1)},\mathbf{H}_k^{(2)}$ by online and target encoders. Then through a readout function (\eg, mean pooling), we obtain global representations $\boldsymbol{g}_k$ as:
\begin{equation}
    \boldsymbol{g}_k=\frac{1}{N_k}\sum_i^{N_k} \mathbf{H}_k(i),
\end{equation}
where $\mathbf{H}_k(i)$ represents the $i$-th node representation in graph $k$. Following that, a projection head is added on top of the encoder to map augmented representations to another latent space where the contrastive loss is calculated. Finally, we will contrast these representations $\boldsymbol{z}$ through \emph{NT-Xent} (the normalized temperature-scaled cross-entropy loss~\cite{chen2020simple}). The sample with its augmented view is considered as a positive pair, and others are considered as negative pairs. The formula of contrastive loss follows:
\begin{equation}
\ell_{i}=-\log \frac{\exp \left(\operatorname{sim}\left(\boldsymbol{z}_i^{(1)}, \boldsymbol{z}_i^{(2)}\right) / \tau\right)}{\sum_{k=1}^B \exp \left(\operatorname{sim}\left(\boldsymbol{z}_i^{(1)}, \boldsymbol{z}_k^{(2)}\right) / \tau\right)},
\end{equation}
where $\operatorname{sim}(\boldsymbol{u}, \boldsymbol{v})=\boldsymbol{u}^{\top} \boldsymbol{v} /\|\boldsymbol{u}\|\|\boldsymbol{v}\|$ computes the similarity score between $\boldsymbol{u}$ and $\boldsymbol{v}$, $\tau$ is temperature parameter, and $B$ denotes the size of mini-batch.

\paragraph{How to Integrate Local And Global Branches?}
Due to consideration of efficiency and performance, the small batch sizes (\eg, 8, 16) are used for training local branch, which is insufficient for effective contrastive learning~\cite{chen2020simple}. Thus, direct integration of these branches is not feasible. To address this, we employ a dynamic queue $\mathbf{Q}$ to incorporate more negative samples and enhance the integration of local and global branches.
This dynamic queue holds the first in first out property and we use the target representations $\mathbf{H}^{(2)}$ to update this dynamic queue $\mathbf{Q}$. In order to keep the consistency of representations in the dynamic queue to the utmost, we use exponential moving average to update the target encoder $\mathcal{E}_t$ and projection head $f_t$.
Formally, the parameters in $\mathcal{E}_t$ is updated by $\theta_t \leftarrow \mu \theta_t+(1-\mu) \theta_o$. Here $m \in [0,1)$ is a momentum coefficient that controls the smoothness of evolving target parameters, we use a relatively large momentum (\eg, 0.999) in our experiments empirically. 
With dynamic queue $\mathbf{Q}$, the loss in global branch:

\begin{equation}
\ell_{i}=-\log \frac{e^{\left(\operatorname{sim}\left(\boldsymbol{z}_i^{(1)}, \boldsymbol{z}_i^{(2)}\right) / \tau\right)}}{\sum_{k=1}^B e^{\left(\operatorname{sim}\left(\boldsymbol{z}_i^{(1)}, \boldsymbol{z}_k^{(2)}\right) / \tau\right)} + \sum_{j=1}^Q e^{\left(\operatorname{sim}\left(\boldsymbol{z}_i^{(1)}, \boldsymbol{q}_j\right) / \tau\right)}}
\label{equ:contra}
\end{equation}
where $Q$ denotes the size of the dynamic queue. $\mathbf{q}$ represents the sampled representations in the dynamic queue. The total loss of the global branch is averaged over batch samples (\ie, $\mathcal{L}_{\text{global}}=\frac{1}{B}\sum_i^{B} \ell_{i}$). 

In summary, the overall pre-training loss is defined as
\begin{equation}
    \mathcal{L}_{\text{pre}} = \lambda_{\text{pre}}\cdot\mathcal{L}_{\text{local}} + \left(1-\lambda_{\text{pre}}\right)\cdot\mathcal{L}_{\text{global}},
\label{equ:pre}
\end{equation}
where $\lambda_{\text{pre}} \in [0,1]$ controls the weight of local loss. For most cases, we set $\lambda_{\text{pre}}$ as 0.5 which means local and global loss contributes equally.

\subsection{Verbalizer-free Graph Prompt Tuning}
After obtaining the pre-trained model from \our~, we propose a novel graph prompt tuning technique that mitigates the representation gap between pre-trained model and downstream tasks. 
Through \emph{prompt addition}, we reformulate the downstream task in the same format as the pre-training task. And with the design of verbalizer-free \emph{prompt answer}, we get rid of the verbalizer which is hard to design in graph domain. 
The process of graph prompt tuning is shown in Figure~\ref{fig:prompt}. And Algorithm~\ref{alg:prompt} in Appendix A offers procedures.

\begin{figure}[htp]
    \centering
    \includegraphics[scale=0.5]{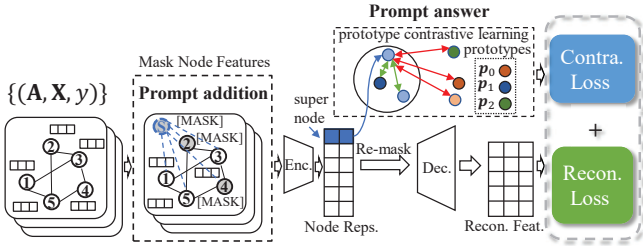}
    \vspace{-2em}
    \caption{The framework of verbalizer-free prompt tuning. }
    \label{fig:prompt}
    \vspace{-1em}
\end{figure}

\paragraph{Prompt Addition: Re-formulate downstream task} 
Prompt-based learning methods mitigate the representation gap by reformulating the downstream task to the same format as the pre-training task, however, it is non-trivial to do so in graph domain due to the inherent abstraction of graph data. 
To solve the problem, we introduce a masked super node, which connects to all nodes in the graph and therefore has a global receptive field.
Thus the representation of the super node can be seen as the representation of the whole graph. 
To reconstruct the features of the masked super node, we transform the original classification task into masked feature reconstruction, which corresponds to the task during pre-training. This idea is motivated by the prompt tuning in NLP domain, which adds a template with a task slot (masked word) for input sentence and predicts the masked word of the slot~\cite{prompt-survey}, remaining different to previous graph prompt works~\cite{gppt}.

\paragraph{Prompt Answer: Verbalizer-free Class Mapping}
We have reformulated the downstream task as masked node feature reconstruction. Since there are no semantic and representative input features related to labels for the masked super nodes, it is hard to use \emph{verbalizer} to establish mappings between reconstructed features and their semantic labels. 

In this work, we get rid of the verbalizer by introducing the supervised prototypical contrastive learning (SPCL)~\cite{li2020prototypical,cui2022prototypical} for class mapping. Specifically, the prototypes represent essential features corresponding to labels. As depicted in Figure~\ref{fig:prompt}, we will obtain class prototypes $\boldsymbol{p}_c$ by SPCL and use these prototypes as semantic tokens related to labels. The supervised prototypical contrastive loss is given by:

\begin{align}
\mathcal{L}_{\text {proto}}=
&\frac{-1}{C^2 B^2} \sum_c \sum_{i, j} \log \frac{\exp\left( \operatorname{sim}\left(\boldsymbol{z}_i^c, \boldsymbol{z}_j^c\right)/ \tau \right)}{\sum_{c^\prime, j^{\prime}} \exp \left(\operatorname{sim}\left(\boldsymbol{z}_i^c, \boldsymbol{z}_{j^{\prime}}^{c^\prime}\right)/ \tau \right)}\notag \\
&+ \frac{-1}{C^2 B} \sum_{i, c} \log \frac{\exp \left(\operatorname{sim}\left(\boldsymbol{z}_i^c, \boldsymbol{p}_c\right)/\tau\right)}{\sum_{c^\prime} \exp\left( \operatorname{sim}\left(\boldsymbol{z}_i^c, \boldsymbol{p}_{c^\prime}\right)/\tau\right)},
\label{equ:proto}
\end{align}
where the first part is instance-instance loss which draws intra-class pairs close and pushes inter-class pairs away in the representation space. And the second part represents instance-prototype loss which makes the similarity scores between instances of class $c$ and prototype $\boldsymbol{p}_c$ larger than other prototypes. Class prototypes $\boldsymbol{p}_c$ are learnable vectors, which are updated by the second part loss. These two parts both use \emph{NT-Xent} which is the same as the global branch.

\paragraph{Prompt Tuning}
Through prompt engineering (\ie, prompt addition and prompt answer), we reformulate the downstream task in the same format as pre-training method, which means the original classification task is transformed into reconstructing the super node's representations. 
In order to keep the training objective consistent with pre-training and avoid catastrophic forgetting of the pre-trained knowledge, we use $\mathcal{L}_{\text{local}}$ as an auxiliary loss with a low masking rate during prompt tuning. 
The overall loss of prompt tuning is defined as:
\begin{equation}
    \mathcal{L}_{\text{prompt}} = \lambda_{\text{prompt}}\cdot\mathcal{L}_{\text{local}} + \left(1-\lambda_{\text{prompt}}\right)\cdot\mathcal{L}_{\text{proto}},
\end{equation}
where $\lambda_{\text{prompt}} \in [0,1]$ controls the  contribution of each component loss. We set $\lambda_{\text{prompt}}$ as 0.1 for all datasets to focus on learning class prototypes used for classification.

In this way, we implement verbalize-free graph prompt tuning. This loss holds a similar format with $\mathcal{L}_{\text{pre}}$. 

\subsection{Prediction}
Similar to prompt tuning, we add a masked super node in the original graph and hope to reconstruct its corresponding class prototype. Specifically, by comparing the representation of the super node $\boldsymbol{g}_{\text{sup}}$ with each class prototype $\boldsymbol{p}_c$, we can get the predicted class probability:
\begin{equation}
    P(y_i|\boldsymbol{g})=\frac{\operatorname{exp}\left(\operatorname{sim}\left(\boldsymbol{g}_{\text{sup}}, \boldsymbol{p}_i\right)\right)}{\sum_{j=1}^C \operatorname{exp}\left(\operatorname{sim}\left(\boldsymbol{g}_{\text{sup}}, \boldsymbol{p}_j\right)\right)}.
\end{equation}

We will choose the highest score as our predicted class:

\begin{equation}
    \tilde{y} = \operatorname{argmax}_k P(y_k|\boldsymbol{g}_{\text{sup}}).
\end{equation}

%% file: sections/experiments.tex
\section{Experiments}

In this section, we will introduce the datasets and experimental setups of graph classification that we used.
Then we evaluate the performance of the proposed self-supervised pre-training method \our. 
Thirdly, we prove the effectiveness of our prompt tuning framework \ourpts compared with standard fine-tuning and other graph prompt methods under semi-supervised and few-shot settings. After that, we conduct ablation study to prove the necessity of the design.
\textbf{Additionally, we perform sensitivity analysis on crucial hyperparameters (\eg, dynamic queue size $Q$ and loss coefficient $\lambda$) in Appendix C. We substantiate the robustness and generality of our pre-training method with supplementary experiments (\eg, molecule property prediction (over 2 million graphs) and node classification) in Appendix C, showcasing SOTA performance. Furthermore, we empirically validate the effectiveness of our prompt design in few-shot node-level tasks within Appendix C. Moreover, we assess parameter efficiency among various prompt methods to show \ourpt's efficiency in Appendix D.}

\subsection{Datasets}
We perform experiments of graph-level tasks on widely used 12 datasets from TUDataset~\cite{tudataset}. The statistics of the used datasets can be found in Table~\ref{tab:graph_cla}. They can be classified into two categories: biological and social networks. More details of these datasets are in Appendix B.


\begin{table}[htp]
\centering
\scalebox{0.8}{
\begin{tabular}{l|c|c|c|c}
\hline
\hline
Datasets & type     & \# graphs & Avg \# nodes & Avg \# edges \\
\hline
\hline
PROTEINS & Biological   & 1113      & 39.06        & 72.82        \\
DD       & Biological   & 1178      & 284.32       & 715.66       \\
MUTAG    & Biological   & 188       & 17.93        & 19.79        \\
NCI1     & Biological   & 4110      & 29.87        & 32.30        \\ 
NCI-H23  & Biological   & 40353     & 26.07        & 28.10        \\ 
P388     & Biological   & 41472     & 22.11	      & 23.56        \\
MOLT-4   & Biological   & 39765     & 26.10	      & 28.14        \\\hline
IMDB-B   & Social       & 1000      & 19.77        & 96.53        \\
IMDB-M   & Social       & 1500      & 13.00        & 65.94        \\
COLLAB   & Social       & 5000      & 74.49        & 2457.78      \\
REDDIT-B & Social       & 2000      & 508.52       & 594.87       \\ 
REDDIT-M12K & Social    & 11929     & 391.41       & 456.89       \\
\hline
\end{tabular}}
\vspace{-0.5em}
\caption{Statistics of graph classification datasets.}
\vspace{-1.5em}
\label{tab:graph_cla}
\end{table}

\subsection{Evaluation of Proposed Pre-trained Method} 
In this section, we will evaluate our pre-training method \ours under unsupervised setting.

\paragraph{Baselines} Our baselines mainly consist of three categories: supervised methods, graph kernel methods, and other unsupervised methods. Specifically, we compare with three supervised baselines: GCN~\cite{kipf2017semi}, GIN~\cite{xu2018how} and DiffPool~\cite{diffpool}. The SOTA graph kernel methods include graphlet kernel (GL)~\cite{gl}, Weisfeiler-Lehman sub-tree kernel (WL)~\cite{wl} and deep graph kernel (DGK)~\cite{dgk}. We also compare with unsupervised representation learning methods including sub2vec~\cite{sub2vec}, graph2vec~\cite{graph2vec}, EdgePred~\cite{hu2020strategies}, InfoGraph~\cite{infograph}, GraphCL~\cite{you2020graph}, JOAO~\cite{you2021graph}, SimGRACE~\cite{xia2022simgrace}, MVGRL~\cite{mvgrl}, InfoGCL~\cite{infogcl} and GraphMAE~\cite{gmae}. The introduction of baselines can be found in Appendix B.

\paragraph{Experiment Setup} The quality of the pre-trained graph encoder is then evaluated by the linear separability of the final representations. Namely, an additional trainable linear classifier is built on top of the frozen encoder following \cite{gmae,you2020graph}. We adopt GIN~\cite{xu2018how} as our encoder and decoder with the default setting in \cite{gmae}. The complete hyper-parameters and more details are listed in Appendix B.

\paragraph{Analysis} 
\begin{table*}[htp]
\centering
\scalebox{0.85}{
\begin{tabular}{c|l|cccc|cccc}
\hline
\hline
                                 & Methods                       & PROTEINS   & DD         & NCI1       & MUTAG      & IMDB-B     & IMDB-M     & COLLAB     & REDDIT-B   \\ 
\hline
\hline
\multirow{3}{*}{Supervised}      
                                 & GCN       & 74.9±3.3   & 75.9±2.5   & 80.2±2.0   & 85.6±5.8   & 70.4±3.4   & 51.9±3.8   & 79.0±1.8   & -                        \\
                                 & GIN        & 76.2±2.8   & 75.3±2.9   & 82.7±1.7   & 89.4±5.6   & 75.1±5.1   & 52.3±2.8   & 80.2±1.9   & 92.4±2.5                 \\
                                 & DiffPool      & 75.1±3.5   & -          & -          & 85.0±10.3  & 72.6±3.9   & -          & 78.9±2.3   & 92.1±2.6                 \\ \hline
\multirow{3}{*}{Graph Kernels}   & GL                 & 71.67±0.55 & 72.54±3.83 & -          & 81.66±2.11 & 65.87±0.98 & 43.89±0.38 & 56.30±0.60 & 77.34±0.18               \\
                                 & WL                  & 72.92±0.56 & 79.78±0.36 & 80.01±0.50 & 80.72±3.00 & 72.30±3.44 & 46.95±0.46 & 69.30±3.44 & 68.82±0.41               \\
                                 & DGK              & 73.30±0.82 & 73.50±1.01 & 80.31±0.46 & 87.44±2.72 & 66.96±0.56 & 44.55±0.52 & 64.66±0.50 & 78.04±0.39               \\ \hline
\multirow{10}{*}{Self-supervised} & sub2vec      & 53.03±5.55 & 54.33±2.44 & 52.89±1.61 &61.05±15.79 & 55.26±1.54 & 36.67±0.83 & 55.26±1.54 & 71.48±0.41              \\
                                 & graph2vec    & 73.30±2.05 & 70.32±2.32 & 73.22±1.81 & 83.15±9.25 & 71.10±0.54 & 50.44±0.87 & 71.10±0.54 & 75.78±1.03             \\
                                 & EdgePred  & 73.12±1.54 & 72.34±1.04 & 74.41±1.50 & 84.49±1.56 & 68.48±1.11 & 44.83±0.65 & 64.80±1.16 & 84.48±0.68            \\
                                 & Infograph    & 74.44±0.31 & 72.85±1.78 & 76.20±1.06 & 89.01±1.13 & 73.03±0.87 & 49.69±0.53 & 70.65±1.13 & 82.50±1.42               \\
                                 & GraphCL   & 74.39±0.45 & 78.62±0.40 & 77.87±0.41 & 86.80±1.34 & 71.14±0.44 & 48.58±0.67 & 71.36±1.15 & 89.53±0.84                \\
                                 & JOAO      & 74.55±0.41 & 77.32±0.54 & 78.07±0.47 & 87.35±1.02 & 70.21±3.08 & 49.20±0.77 & 69.50±0.36 & 85.29±1.35   \\
                                 & SimGRACE  & 75.35±0.09 & 77.44±1.11 & 79.12±0.44 & 89.01±1.31 & 71.30±0.77 & -          & 71.72±0.82 & 89.51±0.89   \\
                                 & MVGRL            & -          & -          & -          & 89.70±1.10 & 74.20±0.70 & 51.20±0.50 & -          & 84.50±0.60                \\
                                 & InfoGCL        & -          & -          & 80.20±0.60 & \textbf{91.20±1.30}& 75.10±0.90 & 51.40±0.80 & 80.00±1.30 & -                \\ 
                                 & GraphMAE          & 75.30±0.39 & 79.42±0.42 & 80.40±0.30 & 88.19±1.26 & 75.52±0.66 & 51.63±0.52 & 80.32±0.46 & 88.01±0.19               \\  \cline{2-10}
                                 & \ours      & \textbf{76.55±0.19} & \textbf{80.54±0.65} & \textbf{80.91±0.42} & 88.83±1.44 & \textbf{75.88±0.47} & \textbf{52.84±0.26} & \textbf{80.80±0.23} & \textbf{89.97±0.48}           \\ \hline
\end{tabular}}
\vspace{-0.5em}
\caption{Graph classification accuracies of supervised, kernel and unsupervised methods on small-scale datasets. The reported results of baselines are from previous papers if available.}
\label{tab:unsupervised}
\vspace{-1em}
\end{table*}

\begin{table}[htp]
    \centering
    \scalebox{0.8}{
    \begin{tabular}{c|cccc}
    \hline
    \hline
                 &  NCI-H23   & MOLT-4     & P388       & RDT-M12K         \\ \hline \hline
            GIN  & 78.41±0.36 & 72.40±1.07 & 82.63±0.90 & 35.01±1.48             \\ \hline     
        EdgePred & 72.49±0.72 & 70.53±1.98 & 72.28±0.97 & 28.65±1.48            \\
        Infograph& 78.69±0.69 & 71.29±0.53 & 78.00±1.38 & 33.14±0.74            \\
        GraphCL  & 76.37±1.08 & 73.26±1.14 & 79.92±1.21 & 33.79±2.47        \\
        JOAO     & 76.69±1.13 & 72.49±0.74 & 79.86±1.96 & 35.76±1.42            \\
        MVGRL    & 78.08±0.46 & 74.63±0.32 & 80.20±0.89 & 32.21±1.35           \\
        GraphMAE & 77.09±0.33 & 73.91±0.51 & 80.55±0.48 & 33.77±1.35       \\ \hline
        SGL      & \textbf{79.98±0.53} & \textbf{75.10±0.52} & \textbf{81.75±0.47} & \textbf{36.24±1.83} \\ \hline
    \end{tabular}}
    \caption{Results of supervised GIN and self-supervised methods on larger scale datasets. The baseline results are obtained using their official implementations.}
    \vspace{-1em}
    \label{tab:larger}
\end{table}

The results are listed in Table \ref{tab:unsupervised} and Table \ref{tab:larger}, from which we can draw the conclusions:

\ours outperforms kernel methods on all datasets by a large margin (\eg, 3.2\% absolute improvement over the SOTA graph kernel method DGK on PROTEINS). And our method even outperforms the best supervised model (\ie, GIN) on eight out of twelve datasets, which closes the gap between unsupervised methods and supervised methods.

Compared with other unsupervised methods, \ours achieves SOTA results except on MUTAG.
We suspect this dataset is too small to inspire the full potential of our pre-training method. On larger scale datasets in Table~\ref{tab:larger}, our method surpasses other methods by considerable margins (\eg, around 1.3\% absolute improvement over other methods on NCI-H23 dataset). The strong results show the superiority of our proposed pre-training method \our.



\subsection{Evaluation of Proposed Prompt-tuning} \label{subsec:prompt}
In this section, we aim to investigate the effectiveness of our proposed prompt method \ourpt. And we will conduct experiments under different settings (\ie, semi-supervised, few-shot settings) to achieve this goal. Our prompt method can be applied to other pre-training methods with a little modifications which can be found in Appendix C.

\begin{table*}[htp]
\centering
\scalebox{0.9}{
\begin{tabular}{l|l|ccccc|cc|c}
\hline
\hline
                               & 10\%L.R.             & PROTEINS   & DD         & NCI1       & MUTAG      & NCI-H23    & IMDB-B     & IMDB-M      &Imp$\left(\%\right)$\\ 
\hline
\hline
\multirow{5}{*}{FT}   & No pre-train.                 & 67.98±0.41 & 68.32±0.39 & 64.00±0.22 & 64.19±0.53 & 55.14±1.29 & 69.10±0.28 & 43.25±0.34  &$\;\;\,$0.00\\ \cline{2-9} 
              & EdgePred                              & 68.12±0.93 & 66.77±0.68 & 63.31±0.72 & 67.72±0.33 & 63.44±1.32 & 67.24±1.63 & 41.57±0.38  &$+$0.88      \\
                 & GraphCL                            & 68.58±0.84 & 68.60±0.49 & 68.08±0.48 & 64.76±0.65 & 65.77±0.52 & 69.95±1.83 & 43.39±0.62  &$+$2.45\\
                         & GraphMAE                   & 68.79±0.77 & 68.70±0.31 & 65.19±0.48 & 71.16±0.83 & 64.25±1.44 & 69.46±0.33 & 44.09±0.49  &$+$2.81\\
                        &$\text{\our}^{\star}$        & 69.41±1.05 & 68.81±0.40 & 66.51±0.28 & 72.51±1.83 & 64.18±1.29 & 69.62±0.24 & 44.69±0.56  &$+$3.39\\ 
\hline
\multirow{7}{*}{PT} & GPPT                            & 68.26±0.87 & 66.53±0.85 & 62.85±1.20 & 71.13±1.81 & 53.44±0.81 & 66.04±0.82 & 38.64±1.40  &$-$0.73 \\
                      & $\text{GPPT}^{\star}$         & 65.28±1.99 & 66.04±0.72 & 66.24±0.51 & 71.29±2.07 & 51.86±0.70 & 69.52±0.55 & 43.56±0.56  &$+$0.26 \\
           &GraphPrompt                               & 71.00±0.46 & 65.24±1.26 & 60.55±2.17 & 71.78±0.75 & 55.39±0.85 & 62.60±2.32 & 40.33±0.60  &$-$0.72  \\
           &$\text{GraphPrompt}^\star$                & 71.26±0.43 & 67.07±0.50 & 64.47±0.27 & 70.42±0.54 & 60.61±2.51 & 69.88±0.23 & 43.44±0.31  &$+$2.16  \\ \cline{2-9} 
           &$\text{GPF}^\star$                        & 67.19±0.75 & 67.97±0.50 & 67.37±0.55 & 71.85±1.04 & 63.24±1.51 & 69.60±0.43 & 43.44±0.43  &$+$2.67  \\
           &$\text{ProG}^\star$                       & 66.88±0.90 & 67.76±0.80 & 64.59±0.40 & 69.16±1.52 & 68.33±0.60 & 70.18±0.54 & 44.23±0.53  &$+$2.73  \\ \cline{2-9} 

                    &$\text{\ourpt}^{\star}$& \textbf{72.94±0.24} & \textbf{75.37±0.16} & \textbf{68.80±0.18} & \textbf{80.07±1.37} & \textbf{69.71±0.15} & \textbf{70.40±0.77} & \textbf{45.28±0.41} &\textbf{$+$7.31} \\
\hline
\end{tabular}}
\vspace{-0.5em}
\caption{Fine-tuning (FT) and prompt tuning (PT) results under semi-supervised setting. 10\%L.R. denotes 10\% label rate of training data. `No pre-train.' means GIN training from scratch. Imp(\%) represents the average improvement of each method over No pre-train. $\star$ means that we use \ours as their pre-trained models.}
\label{tab:prompt}
\vspace{-0.5em}
\end{table*}

\subsection{Semi-supervised Setting} \label{sec:semi}
Experiments follow a semi-supervised setting with pre-training \& fine-tuning~\cite{chen2020simple,you2020graph}. We don't fix the pre-trained model and tune all parameters for downstream tasks. In more limited source scenarios like the next section (few-shot setting), we will freeze the pre-trained model and only train additional parameters for downstream tasks (\eg, classifier).
\paragraph{Baselines} To thoroughly investigate the effectiveness of the proposed \ourpt, we compare it with methods of different training strategies.

Firstly all the baselines (except `No pre-train.') are pre-trained with pretext tasks. Then for fine-tuning methods, we obtain the pre-trained models in advance and fine-tune them with a linear classifier on downstream labeled data. 

For GPPT, EdgePred serves as the pre-training method, with downstream tasks reformulated into edge prediction tasks following their prompt design.
GraphPrompt, akin to GPPT, employs EdgePred for pre-training and adapts downstream tasks into edge prediction, employing a simplified prompt template featuring weighted summation readout.
$\text{GPF}^\star$ employs \ours as the pre-trained model, integrating a learnable graph prompt feature onto node attributes. In the case of $\text{ProG}^\star$, \ours is also the pre-trained model, incorporating a learnable prompt graph into the original graph.
In the context of \ourpt, \ours serves as the pre-trained model, and we reframe downstream tasks in a similar format to the pre-training task.
Furthermore, in order to demonstrate the superiority our prompt method is not only dependent on a superior pre-train model, we replace the EdgePred with \ours in GPPT, GraphPrompt which are coined as $\text{GPPT}^\star$ and $\text{GraphPrompt}^\star$ respectively. We use grid search on important hyper-parameters to get the best performance.

\paragraph{Experimental Setup} 
We ensure a fair comparison by using the same model configuration for all methods. 
The detailed settings and hyper-parameters are in Appendix B. 

\paragraph{Analysis} Table~\ref{tab:prompt} summarizes the results of different training strategies, and we can get the following information:

For fine-tuning-based methods, the order of performance follows ``EdgePred $<$ GraphCL $<$ GraphMAE $<$ \our'', which is similar to the results in previous unsupervised learning. Our method \ours surpasses other fine-tuning-based methods, which again testifies the effectiveness of our pre-training method \our. It is worth noting that EdgePred even cannot outperform `No pre-train.' on some datasets, which indicates EdgePred method triggers negative transfer.

\ourpts outperforms all fine-tuning-based methods and surpasses \ours by a large margin (around 3\% average improvement on all datasets), which proves the existence of representation gap between pre-training and downstream tasks and the urgency of minimizing such gap. 
It also shows that \ourpts outperforms GPPT and GraphPrompt by a large margin (around 6\% average improvement on all datasets). The reason GPPT and GraphPrompt perform poorly is that they only utilize limited learned knowledge of the pre-trained model. And \ourpts surpasses $\text{GPF}^\star$ and $\text{ProG}^\star$ by a considerable margin (around 5\% average improvement on all datasets), which proves the effectiveness of unifying the training objectives.

We can find $\text{GPPT}^\star$, $\text{GPF}^\star$, $\text{ProG}^\star$ and $\text{GraphPrompt}^\star$ even perform worse than SGL due to catastrophic forgetting from inconsistent objectives on some datasets (\eg, For DD dataset, $\text{GPPT}^\star$, $\text{GPF}^\star$ and $\text{GraphPrompt}^\star$ lag 2.8\%, 0.8\%, 6.2\% and 1.8\% behind $\text{SGL}^\star$ respectively). This experiment serves as proof that the effectiveness of our prompt tuning method is not dependent on a superior pre-train model, what matters is to establish consistent training objectives that align with both pre-train and downstream tasks.


\begin{table*}[htp]
\centering
\scalebox{0.75}{
\begin{tabular}{l|l|cc|cc|cc|cc|cc}
\hline
\hline
                            &\multirow{2}{*}{Few-shot}       &\multicolumn{2}{c|}{PROTEINS}  & \multicolumn{2}{c|}{DD}     &\multicolumn{2}{c|}{NCI1}  &\multicolumn{2}{c|}{MUATG}   &\multicolumn{2}{c}{NCI-H23}\\
                            &       &1-shot      & 3-shot           &1-shot      & 3-shot         &1-shot      & 3-shot       &1-shot      & 3-shot        & 1-shot      & 3-shot       \\
\hline
\hline
\multirow{4}{*}{FT}
&EdgePred      & 57.25±2.80 & 58.36±2.06      & 50.88±2.85 & 52.02±1.21      & 50.75±0.30 & 50.94±0.56   & 57.01±4.83 & 58.71±4.04    & 51.94±1.85  & 52.45±2.35 \\
&GraphCL       & 59.40±1.87 & 59.91±2.88      & 52.30±2.51 & 53.39±2.87      & 51.51±0.28 & 51.72±0.55   & 57.92±3.43 & 62.39±1.53    & 52.28±3.13  & 54.85±2.14 \\
&GraphMAE      & 58.46±3.75 & 59.29±1.62      & 51.31±1.63 & 54.14±1.15      & 51.45±1.45 & 52.67±1.86   & 57.67±4.11 & 60.92±3.54    & 50.40±1.58  & 53.72±1.61 \\
&$\text{SGL}^\star$                 & 58.76±1.02 & 59.73±2.65      & 52.87±4.15 & 54.51±2.91      & 51.88±1.65 & 53.22±2.11   & 58.68±2.28 & 62.47±2.51    & 52.06±2.69  & 55.67±1.08 \\
\hline
\multirow{7}{*}{PT}
&GPPT          & 58.10±1.11 & 58.63±0.81      & 50.45±1.49 & 51.50±2.04      & 51.50±0.81 & 51.25±0.47   & 63.23±4.35 & 66.06±2.55    & 51.55±1.19  & 51.84±1.14\\
&$\text{GPPT}^\star$   & 58.64±2.43 & 60.02±1.66      & 54.47±3.49 & 56.92±2.20      & 51.29±2.16 & 52.10±2.37   & 63.13±3.53 & 65.86±1.65    & 49.88±2.20  & 50.03±1.96 \\
&GraphPrompt  & 58.94±2.37 & 62.01±1.47      & 53.62±1.09 & 53.38±1.37      & 51.49±0.38 & 51.64±0.58   & 67.62±1.93 & 70.03±2.18    & 55.03±0.75  & 56.97±1.25 \\
&$\text{GraphPrompt}^\star$         & 59.95±2.71 & 62.93±1.37      & 55.03±2.40 & 58.21±0.95      & 52.65±1.06 & 52.50±1.21   & 68.22±2.86 & 71.40±3.12    & 52.90±2.47  & 52.57±1.89 \\ \cline{2-12} 
&$\text{GPF}^\star$                 & 57.76±2.07 & 59.32±2.43      & 53.58±2.99 & 54.24±2.15      & 51.68±0.40 & 51.91±0.88   & 60.15±4.98 & 61.93±3.03    & 51.19±2.93  & 54.20±3.77 \\
&$\text{ProG}^\star$                & 57.99±1.47 & 60.99±0.96      & 52.75±3.27 & 54.32±1.57      & 51.65±1.04 & 52.03±0.35   & 61.37±2.78 & 62.42±1.70    & 50.48±1.27  & 52.66±1.94 \\ \cline{2-12} 
&$\text{SGL-PT}^{\star}$            & \textbf{61.02±2.63} & \textbf{64.47±2.10}      & \textbf{57.15±1.92} & \textbf{61.12±2.42}      & \textbf{53.08±2.06} & \textbf{55.24±1.28}   & \textbf{72.88±5.20} & \textbf{78.60±1.81}  & \textbf{55.61±1.74} & \textbf{58.76±2.09}   \\
\hline
\end{tabular}}
\vspace{-0.5em}
\caption{Fine-tuning\&prompt tuning results under few-shot setting. $\star$ means that we use \ours as their pre-trained models.}
\label{tab:fewshot}
\vspace{-0.2em}
\end{table*}

\begin{table*}[ht]
\centering
\scalebox{0.87}{
\begin{tabular}{l|cccc|cccc}
\hline
\hline
                  & PROTEINS   & DD         & NCI1       & MUTAG      & IMDB-B     & IMDB-M     & COLLAB     & REDDIT-B   \\
\hline
\hline
Full              & \textbf{76.55±0.19} & \textbf{80.54±0.65} & \textbf{80.91±0.42} & \textbf{88.83±1.44} & \textbf{75.88±0.47} & \textbf{52.84±0.26} & \textbf{80.80±0.23} & \textbf{89.97±0.48} \\
w/o local branch  & 74.68±0.50 & 79.67±0.46 & 79.12±0.35 & 85.22±1.25 & 74.42±0.16 & 50.59±0.32 & 78.19±0.12 & 86.67±0.37 \\
w/o global branch & 75.18±0.40 & 79.54±0.54 & 80.18±0.28 & 86.81±2.12 & 75.24±0.55 & 51.63±0.52 & 80.33±0.38 & 87.83±0.11 \\
w/o dynamic queue & 74.85±0.43 & 78.50±0.46 & 80.37±0.15 & 85.23±1.13 & 74.89±0.38 & 51.06±0.71 & 80.51±0.29 & 86.70±0.50 \\
\hline
\end{tabular}}
\vspace{-0.5em}
\caption{Ablation studies for \ours by masking local branch, global branch and dynamic queue under the unsupervised setting.}
\label{tab:aba}
\vspace{-1em}
\end{table*}

\subsection{Few-shot Setting} \label{sec:fewshot}
In many real-world scenarios, it is challenging to collect and label a large amount of data. Few-shot learning is a well-known case of low-resource scenarios. We conduct experiments in such a setting to prove the effectiveness of our method in low-resource scenarios. Experiments on more datasets can be found in Appendix C.


\paragraph{Experimental Setup} 
In this section, we evaluate different training strategies with more limited supervision in a few-shot setting. This entails having only a small number of labeled graphs per class, denoted as $k$-shot classification. We perform experiments with 1-shot and 3-shot graph classification to evaluate all methods. The model setup remains consistent with the previous section. Additionally, for prompt methods, we freeze the pre-trained model and exclusively train supplementary parameters for downstream tasks.

\paragraph{Analysis} Table~\ref{tab:fewshot} summarizes the results of different training strategies, and we can obtain the following results:


In few-shot setting, prompt methods can achieve better performance than standard fine-tuning-based methods (\eg, GPPT outperforms EdgePred, \ourpts outperforms \our.) which proves the essential of unifying the pre-training and downstream tasks. 

And \ourpts can still outperform other methods which proves the effectiveness of our prompt design even in low-resource scenarios. (\eg, \ourpts surpasses other methods by over 7 \% absolute improvement on the MUTAG of 5-shot.)

Even with a stronger pre-trained model, $\text{GPPT}^{\star}$, $\text{GPF}^{\star}$, $\text{ProG}^{\star}$ and $\text{GraphPrompt}^{\star}$ still perform moderately on most datasets, highlighting the importance of unifying the training objectives of pre-training and downstream tasks to fully exploit learned knowledge. These findings underscore the necessity to solve the challenges we proposed simultaneously.


\subsection{Ablation Study}
To prove the effectiveness of the design of our pre-training method \our, we conduct ablation experiments that mask different components under the same model configuration. \ours contains two branches (\ie, \emph{local} and \emph{global}), and we mask them separately. `w/o global branch' means that we only use the local branch. `w/o local branch' means that we only use the contrastive learning method. And `w/o dynamic queue' represents that we do not use the dynamic queue to provide adequate negative samples. 
From Table~\ref{tab:aba}, the results of `w/o dynamic queue' lags far behind \our, which demonstrates the dynamic queue is essential to integrate these two branches better. Disjunct single branches also lag behind \our, this means through efficient combination, we acquire complementary strengths of both generative (local branch) and contrastive (global branch) methods.

%% file: sections/conclusion.tex
\section{Conclusion}
In this work, we identify the main challenges for graph prompt tuning. To solve them, we propose a ``Pre-train, prompt and predict'' framework coined as \ourpt. This framework unifies the pre-training and downstream task in the same format and minimizes the training objective gap. Specifically, we design a strong and universal pre-training task that acquires the complementary strengths of generative and contrastive methods. Based on this pre-training method, we design a novel verbalizer-free prompting function to reformulate the downstream task in the same format as our pre-training method. Empirical results show that our pre-training method surpasses other baselines under the unsupervised setting, and our prompt tuning method can greatly facilitate pre-trained models compared to standard fine-tuning methods and other graph prompt tuning techniques.

%% file: sections/appendix.tex
\begin{appendix}

\section*{A. Algorithm}\label{app:alg}
\subsection{Algorithm of SGL}
The overall processes of \ours can be described as Algorithm~\ref{alg:pretrain}. The input graphs will be fed into two branches (\ie, \emph{local} and \emph{global}). In the local branch, we will mask partial node features. Then through the online encoder $\mathcal{E}_o$, we will obtain fragmentary representations. Following \cite{gmae}, we will re-mask node representations. And finally through a decoder, we restore the input features on masked nodes. The reconstructed loss is computed by Equation 1.
Meanwhile, in the global branch, we will augment input graphs and feed them into target encoder $\mathcal{E}_t$, readout function $\mathcal{R}$ and target projection head $f_{t}$ to obtain global representations $\boldsymbol{z}^{(2)}$, and we will make use of node representation in local branch to obtain other view representations $\boldsymbol{z}^{(1)}$ through readout function and online projection head $f_o$. Finally, through Equation 5
, we obtain the global loss in the global branch. And the total loss is computed by a weighted sum of $\mathcal{L}_{\text{local}}, \mathcal{L}_{\text{global}}$. The online encoder, projection head and decoder are updated by gradient descent. But the target encoder and projection head are updated by momentum update.
\begin{algorithm}
\caption{Pretraining method: \our}\label{alg:pretrain}
\KwIn{Online and target encoders $\mathcal{E}_o, \mathcal{E}_t$, online and target projectors $f_o, f_t$, decoder $\mathcal{D}$ and readout function $\mathcal{R}$, 
 feature mask function $\mathcal{M}$ and augmentation function $\mathcal{A}$, unlabeled graph classification dataset $\mathbb{G}$, training epochs $E$.}
\KwOut{Optimal pre-trained encoder $\mathcal{E}_o^{\star}$}

\For{$i\leftarrow 1$ \KwTo $E$}{
\tcp{For Local branch}
$G^{(1)} \leftarrow \mathcal{M}(G)$, $\mathbf{H}^{(1)} \leftarrow \mathcal{E}_o(G^{(1)})$ \;
$\tilde{\mathbf{X}}^{(1)} \leftarrow \mathcal{D}(\mathbf{H}^{(1)})$ \;
Calculate the loss $\mathcal{L}_{\text{local}}$ of local branch according to Equation~\ref{equ:sce}\;
\tcp{For global branch}\
$G^{(2)} \leftarrow \mathcal{A}(G)$, $\mathbf{H}^{(2)} \leftarrow \mathcal{E}_o(G^{(2)})$\;
Summarize graph representation through readout function $\boldsymbol{g}^{(1)} \leftarrow \mathcal{R}(\mathbf{H}^{(1)}), \boldsymbol{g}^{(2)} \leftarrow \mathcal{R}(\mathbf{H}^{(2)})$ \;
$\boldsymbol{z}^{(1)} \leftarrow f_o(\boldsymbol{g}^{(1)})$, $\boldsymbol{z}^{(2)} \leftarrow f_t(\boldsymbol{g}^{(2)})$ \;
Calculate the loss $\mathcal{L}_{\text{global}}$ of global branch according to Equation~\ref{equ:contra}\;
$\mathcal{L}_{\text{pre}} = \lambda_{\text{pre}}\cdot\mathcal{L}_{\text{local}} + \left(1-\lambda_{\text{pre}}\right)\cdot\mathcal{L}_{\text{global}}$ \;
Update parameters of $\mathcal{E}_o$, $\mathcal{D}$ and $f_o$ by applying gradient descent to minimize $\mathcal{L}_{\text{pre}}$ \;
\tcp{Momentum update}\
$\boldsymbol{\theta}_{t}^i \leftarrow m\boldsymbol{\theta}_t^{i-1} + (1-m)\boldsymbol{\theta}_o^i$ \;
Using $\boldsymbol{z}^{(2)}$ to update dynamic queue $\mathbf{Q}$ \;
}
\end{algorithm}

\subsection{Algorithm of Graph Prompt Tuning}
The overall process of graph prompt tuning \ourpts can be concluded in Algorithm~\ref{alg:prompt}. Firstly, we will add a masked super node on each downstream graph and reformulate the downstream task as masked node prediction. We use prototype contrastive to get rid of the verbalizer, and additionally we will mask partial node features and calculate reconstruction loss to avoid catastrophic forgetting. Lastly, the prototype contrastive loss and reconstruction loss are linearly combined through coefficient $\lambda_{\text{prompt}}$.
\begin{algorithm}
\caption{Graph Prompt Tuning}\label{alg:prompt}
\KwIn{Pretrained encoder $\mathcal{E}_o^{\star}$, prompt addition function $\mathcal{P}$, feature mask function $\mathcal{M}$ , labeled downstream dataset $\mathbb{G}$, training epochs $E$}

\For{$i\leftarrow 1$ \KwTo $E$}{
$\hat{G} \leftarrow \mathcal{M} \circ \mathcal{P}(G)$, $\mathbf{H} \leftarrow \mathcal{E}_o^{\star}(\hat{G})$ \;
$\tilde{X} \leftarrow \mathcal{D}(\mathbf{H})$ \;
Calculate the reconstruction loss $\mathcal{L}_{\text{local}}$ according to Equation~\ref{equ:sce}\;
Select the super node representation as graph representation $\mathbf{g} \leftarrow \mathbf{H}[0]$ \;
Calculate the prototype contrastive loss $\mathcal{L}_{\text{proto}}$ according to Equation~\ref{equ:proto}\;
$\mathcal{L}_{\text{prompt}} = \lambda_{\text{prompt}}\cdot\mathcal{L}_{\text{local}} + \left(1-\lambda_{\text{prompt}}\right)\cdot\mathcal{L}_{\text{proto}}$ \;
Update parameters of $\mathcal{E}_o^{\star}$ by applying gradient descent to minimize $\mathcal{L}_{\text{prompt}}$.
}
\end{algorithm}

\section*{B. Experimental Details}\label{app:implement}
\subsection{Datasets}
In this subsection, we give a more detailed description of the datasets used in the main paper. More descriptions can be found in \cite{yan2008mining,dgk}.

\paragraph{For bioinformatics datasets:} 1. MUTAG: This dataset consists of 188 mutagenic aromatic and heteroaromatic nitro compounds. Each compound is represented as a graph, with nodes corresponding to atoms and edges indicating bonds between them. There are 7 discrete labels associated with the mutagenic activity of the compounds.

2. PROTEINS: In this dataset, nodes represent secondary structure elements (SSEs) in protein structures. An edge is established between two nodes if the corresponding SSEs are neighboring in the amino-acid sequence or in 3D space. The dataset contains protein structures with 3 discrete labels, representing helix, sheet, or turn.

3. DD: The dataset comprises 1178 protein structures, each depicted as a graph wherein amino acids serve as nodes, and edges connect two nodes if their spatial separation is within 6 Angstroms. The objective of this predictive endeavor is the categorization of the protein structures as either enzymatic or non-enzymatic entities. Importantly, it's worth noting that nodes are uniformly labeled across all datasets.

4. NCI1: Derived from the National Cancer Institute (NCI), this dataset is a subset of chemoinformatics datasets. It contains chemical compounds that have been screened for their ability to suppress or inhibit the growth of a panel of human tumor cell lines. NCI1 has 37 discrete labels, representing different outcomes of the cell line growth inhibition assay.

5. NCI-H23, MOLT-4 and P388: They offer insights into the biological activities of small molecules, including bioassay records for anticancer screenings across various cancer cell lines, specifically categorized as 'Non-Small Cell Lung', 'Leukemia' and 'Leukemia' respectively.

These datasets offer a diverse range of chemical and biological contexts, providing challenges for graph classification tasks with varying numbers of labels and graph structures.

\paragraph{For social network datasets:}
1. IMDB-BINARY and IMDB-MULTI: These datasets are based on movie collaboration networks. Each graph represents the ego-network of an actor or actress, where nodes correspond to individuals and an edge is present between two nodes if they have collaborated in the same movie. IMDB-BINARY is a binary classification task where the goal is to classify each graph into a specific movie genre. IMDB-MULTI is a multi-class classification task with the objective of classifying the graphs into various movie genres.

2. REDDIT-BINARY and REDDIT-M12K: These datasets capture online discussion threads. Nodes in these graphs represent users participating in discussions, and edges are formed when users interact by responding to each other's comments. REDDIT-BINARY involves binary classification, aiming to classify graphs into specific community or subreddit labels. REDDIT-M12K extends this to multi-class classification consisting of 11 different subreddits, namely, "AskReddit, AdviceAnimals, atheism, aww, IAmA, mildlyinteresting, Showerthoughts, videos, todayilearned, worldnews, TrollXChromosomes." And the goal is to predict which subreddit a given discussion graph belongs to.

3. COLLAB: Derived from scientific collaboration data, this dataset consists of ego-networks representing researchers in different fields. Each graph focuses on researchers from a specific scientific field, such as High Energy Physics, Condensed Matter Physics, or Astrophysics. The classification task involves assigning each graph to the corresponding scientific field.

These datasets provide diverse scenarios for evaluating graph classification algorithms, with each graph representing a unique context of collaboration or interaction, and the classification tasks focusing on predicting genres, communities, or scientific fields.

It has to note that social networks do not contain raw node attributes, we use node degrees as their attributes following \cite{gmae}. As for biological graphs, we use their categorical node attributes.

\subsection{Baselines}
In this part, we will introduce the baselines used in our experiments:
\begin{itemize}
    \item Edge prediction (\textbf{EdgePred})~\cite{hu2020strategies} treat existing links as training signals. Its training objective is binary cross-entropy loss.
    \item Unsupervised And Semi-Supervised Graph-level Representation Learning via Mutual Information Maximization (\textbf{InfoGraph})~\cite{infograph} takes a pair of global representation and patch representation as input and employs a discriminator to determine if they belong to the same graph based on Deep InfoMax~\cite{DIM}. This process generates all possible positive and negative samples in a batch-wise manner.
    \item Graph Contrastive Learning with Augmentations (\textbf{GraphCL})~\cite{you2020graph} proposes various augmentation techniques for graph data and investigates their impacts on different types of datasets. Firstly, the input graph will be fed into two random augmentation functions to generate two graph views, then these augmented graphs will be fed into GNN encoder with readout function to obtain graph representations. Finally, these graph representations will be used to contrast with InfoNCE loss~\cite{cpc}. 
    \item Graph Contrastive Learning Automated (\textbf{JOAO})~\cite{you2021graph} introduces the concept of joint augmentation optimization, which formulates a bi-level optimization problem by simultaneously optimizing the selection of augmentations and the contrastive objective.
    \item A Simple Framework for Graph Contrastive Learning without Data Augmentation (\textbf{SimGRACE})~\cite{xia2022simgrace} eliminates data augmentation while introducing encoder perturbations to generate distinct views for graph contrastive learning.
    \item Contrastive Multi-View Representation Learning on Graphs (\textbf{MVGRL})~\cite{mvgrl} utilizes the information of multi-views for contrasting. Firstly, it will use the edge diffusion function to generate an augmented graph. And asymmetric encoders will be applied on the original graph and diffusion graph to acquire node embeddings. Next, a readout function is employed to derive graph-level representations. Original node representations and augmented graph-level representation are regarded positive pairs. Additionally, the augmented node representations and original graph-level representation are also considered as positive pairs. The negative pairs are constructed through random shuffling.
    \item Information-Aware Graph Contrastive Learning (\textbf{InfoGCL})~\cite{infogcl} suggests minimizing the mutual information between contrastive parts while preserving task-relevant information within both the individual module and the overall framework. This approach aims to minimize information loss during graph representation learning, following the Information Bottleneck principle~\cite{dib}.
    \item Self-Supervised Masked Graph Autoencoders (\textbf{GraphMAE})~\cite{gmae} is a masked autoencoder. It will mask partial input node attributes firstly and then the encoder will compress the masked graph into latent space, finally a decoder aims to reconstruct the masked attributes. 
\end{itemize}

\subsection{Evaluation Protocol}
\paragraph{Unsupervised Representation Learning} 
For small-scale datasets, we follow \cite{gmae,you2020graph} to assess our \ours pre-training method. After pre-training, we keep the model fixed to generate graph-level representations. These representations are then fed into a downstream LIB-SVM~\cite{libsvm} classifier on small-scale datasets. It's important to note that all self-supervised methods are trained using unsupervised data. The evaluation of the pre-trained model involves training only a classifier with supervised data from the downstream task. Reported results include the mean accuracy from 10-fold cross-validation, with standard deviation after 5 runs, on small-scale datasets.

For larger-scale datasets (NCI-H23, MOLT-4, P388, REDDIT-M12K), due to LIB-SVM's convergence issues, we adopt a one-layer MLP as the downstream classifier. Results are reported as the mean performance from 5-fold cross-validation, using accuracy for REDDIT-M12K and ROC-AUC \cite{davis2006relationship} for others.

\paragraph{Semi-supervised Setting} To assess our proposed prompt method, we carry out experiments in both semi-supervised and few-shot settings. In the semi-supervised setup, we fix the label rate at 10\%, indicating that only 10\% of the training data contains labels. We perform parameter tuning for both encoders and additional downstream components (\eg, classifier) to optimize performance on downstream tasks.

\paragraph{Few-shot Setting} In contrast to the previous section, we introduce even scarcer supervised signals in this scenario to simulate low-resource conditions. Specifically, each class comprises only one or three instances for training data, known as 1-shot and 3-shot graph classification. To address concerns of overfitting and parameter efficiency, we solely fine-tune the downstream parameters in the few-shot setting.

\begin{table*}[htp]
\scalebox{0.9}{
\begin{tabular}{c|c|cccc|cccc|c}
\hline
\hline
                                     &                & PROTEINS & DD    & NCI1  & MUTAG & IMDB-B & IMDB-M & COLLAB & REDDIT-B & ZINC \\
\hline
\hline
\multirow{4}{*}{\makecell[c]{Model \\configuration}} 
                                     & hidden\_size   & 512      & 512   & 512   & 32    & 512    & 512    & 256   & 512     &300\\
                                     & num\_layer     & 3        & 2     & 2     & 5     & 2      & 3      & 2      & 2      &5\\
                                     & activation     & prelu    & prelu & prelu & prelu & prelu  & prelu  & relu   & prelu  &relu    \\
                                     & norm           & BN       & BN    & BN    & BN    & BN     & BN     & BN     & LN     &BN  \\
\hline
\multirow{3}{*}{Local branch}        & scaling factor & 1        & 1     & 2     & 2     & 1      & 1      & 1      & 1      &1  \\
                                     & masking rate   & 0.5      & 0.1   & 0.25  & 0.75  & 20     & 0.5    & 0.75   & 0.75   &0.25   \\
                                     & replace rate   & 0.0      & 0.1   & 0.1   & 0.1   & 0.001  & 0.0    & 0.0    & 0.1    &0.0  \\
\hline
\multirow{7}{*}{Global branch}       & $Q$              & 1024     & 1024  & 1024  & 1024  & 1024   & 1024   & 1024  & 1024      &4096 \\
                                     & momentum       & 0.995    & 0.999 & 0.999 & 0.999 & 0.999  & 0.995  & 0.999  & 0.999  &0.999     \\
                                     & tempurate      & 2        & 2     & 2     & 2     & 2      & 2      & 2      & 0.08   &0.05  \\
                                     & feat\_mask1    & 0.4      & 0.1   & 0.0   & 0.2   & 0.2    & 0.0    & 0.2    & 0.3    &0.0 \\
                                     & feat\_mask2    & 0.1      & 0.2   & 0.0   & 0.5   & 0.5    & 0.2    & 0.3    & 0.3    &0.0 \\
                                     & drop\_edge1    & 0.0      & 0     & 0.0   & 0.0   & 0.1    & 0.0    & 0.0    & 0.0    &0.0 \\
                                     & drop\_edge2    & 0.1      & 0     & 0.0   & 0.3   & 0.2    & 0.4    & 0.2    & 0.0    &0.0 \\
\hline
\multirow{6}{*}{Training}            & batch\_size    & 32       & 32    & 16    & 64    & 32     & 32     & 32     & 8      &256  \\
                                     & epochs         & 100      & 80    & 300   & 22    & 60     & 50      & 20     & 120   &100     \\
                                     & learning rate  & 0.00015  & 0.001 & 0.001 & 0.0005& 0.00015& 0.00015& 0.00015& 0.00015&0.001     \\
                                     & weight\_decay  & 0.0      & 0.0   & 0.0   & 0.0   & 0.0    & 0.0    & 0.0    & 0.0    &0.0     \\
                                     & optimizer      & Adam     & Adam  & Adam  & Adam  & Adam   & Adam   & Adam   & Adam   & Adam\\
                                     & scheduler      & False    & True  & True  & False & False  & False   & True  & False  & False  \\
\hline
\end{tabular}}
\caption{Hyper-parameters for pre-training method.}
\label{tab:hyper_pre}
\end{table*}

\begin{table}[htp]
\scalebox{0.8}{
\begin{tabular}{c|c|cccc}
\hline
\hline
                                     &                & NCI-H23 & MOLT-4 & P388  &RDT-M12K\\
\hline
\hline
\multirow{4}{*}{\makecell[c]{Model}} 
                                     & hidden\_size   & 128      & 128   & 512   & 32    \\
                                     & num\_layer     & 3        & 3     & 2     & 5     \\
                                     & activation     & prelu    & prelu & prelu & prelu  \\
                                     & norm           & BN       & BN    & BN    & BN    \\
\hline
\multirow{3}{*}{Local}        & scaling factor & 2        & 2     & 2     & 2     \\
                                     & masking rate   & 0.25     & 0.25  & 0.25  & 0.75   \\
                                     & replace rate   & 0.1      & 0.1   & 0.1   & 0.1   \\
\hline
\multirow{7}{*}{Global}       & $Q$            & 1024     & 1024  & 1024  & 1024   \\
                                     & momentum       & 0.999    & 0.999 & 0.999 & 0.999    \\
                                     & tempurate      & 2        & 2     & 2     & 2       \\
                                     & feat\_mask1    & 0.0      & 0.1   & 0.2   & 0.2   \\
                                     & feat\_mask2    & 0.2      & 0.1   & 0.2   & 0.4   \\
                                     & drop\_edge1    & 0.0      & 0.0   & 0.0   & 0.0   \\
                                     & drop\_edge2    & 0.0      & 0.0   & 0.0   & 0.4   \\
\hline
\multirow{6}{*}{Training}            & batch\_size    & 16       & 32    & 16    & 32      \\
                                     & epochs         & 100      & 100   & 100   & 100      \\
                                     & learning rate  & 0.0001   & 0.0001& 0.0001& 0.00015    \\
                                     & weight\_decay  & 5e-4     & 5e-4  & 5e-4  & 0     \\
                                     & optimizer      & Adam     & Adam  & Adam  & Adam    \\
                                     & scheduler      & True     & True  & True  & True   \\
\hline
\end{tabular}}
\caption{Hyper-parameters for pre-training method.}
\label{tab:hyper_pre2}
\end{table}

\subsection{Hyper-parameters}
All hyper-parameters used in unsupervised learning are listed in Table~\ref{tab:hyper_pre}. The coefficients in Equation~\ref{equ:pre}
is 0.5 for most datasets except COLLAB and NCI1 (0.9). For the local branch, we adopt similar hyper-parameters in \cite{gmae}. And for the global branch, we do not use augmentation in relatively large datasets~(\eg, NCI1 and ZINC). For other small datasets, we use grid search to obtain the augmentation ratios. For the training hyper-parameters (\eg, batch size, epochs and etc), we adopt similar settings in \cite{gmae}.

\begin{table}[htp]
\begin{tabular}{c|ccccc}
\hline
\hline
         & mask.         & epochs & lr            & bs          & optimizer \\
\hline
\hline
PROTEINS & 0.1        & 30     & 0.01          & 32          & Adam      \\
DD       & 0.1        & 50     & 0.0001        & 32          & Adam      \\
NCI1     & 0.1        & 50     & 0.001         & 16          & Adam      \\
MUTAG    & 0.1        & 50     & 0.001         & 64          & Adam      \\ 
NCI-H23  & 0.1        & 10     & 0.001         & 32          & Adam      \\
MOLT-4   & 0.1        & 10     & 0.01          & 32          & Adam      \\
P388     & 0.1        & 10     & 0.001         & 32          & Adam      \\
\hline
IMDB-B   & 0.1        & 20     & 0.001         & 32          & Adam      \\
IMDB-M   & 0.1        & 20     & 0.001         & 32          & Adam      \\
\hline
\end{tabular}
\caption{Hyper-parameters for Prompt Tuning. `mask.' means masking rate, `lr' represents learning rate and `bs' is batch size.}
\label{tab:prompt_hyper}
\end{table}

As for the hyper-parameters in the prompt tuning, Table~\ref{tab:prompt_hyper} gives you the details. During prompt tuning, we still mask partial nodes' features at a low rate mask (10\%) to avoid the catastrophic forgetting of the pre-training knowledge and over-fitting.
For all methods, we use the same hyper-parameters for a fair comparison. You can increase the number of training epochs, the downstream performance may be boosted further. All the pre-trained models are finetuned under the same setting. Some hyper-parameters are searched with grid search. The learning rate is searched in {0.01, 0.001, 0.0001} and the readout function is in {mean, max, sum}. And we use the same batch size.

\subsection{Computer Infrastructures Specifications}
For hardware, most experiments are conducted on a computer server with four GeForce RTX 2080Ti GPUs with 11GB memory and 48 Intel(R) Xeon(R) CPU E5-2678 v3 @ 2.50GHz. Besides, our models are implemented by Pytorch Geometric 2.0.4~\cite{fey2019fast}, DGL 0.9.1~\cite{wang2019dgl} and Pytorch 1.11.0~\cite{paszke2019pytorch}. All the datasets used in our work are available in DGL and PyTorch Geometric libraries. For molecular property prediction, our implementation is based on the code in https://github.com/snap-stanford/pretrain-gnns with Pytorch Geometric 2.0.4 on GeForce RTX 3090.
\end{appendix}

\section*{C. Additional Experiments}
Due to space constraints in the main content, this section will encompass additional experiments involving molecule property prediction and node classification. These endeavors aim to demonstrate the robustness and versatility of our method. Furthermore, we will delve into an analysis of important hyperparameters, specifically focusing on the dynamic queue size ($Q$) and loss coefficient ($\lambda$). Lastly, we will evaluate our graph prompting method on more datasets and apply it to other pre-training methods.

\subsection{Molecule Property Prediction}\label{app:addition}
Besides the single-label graph classification task in the main content, we also evaluate our pre-training method on another graph-level task (\ie, molecular property prediction) to predict chemical molecule properties. In this experiment, a larger molecule dataset is employed to pre-train the model, followed by fine-tuning on smaller downstream datasets. This experiment serves to assess the ability of our pre-training method to generalize across different distributions, showcasing its potential for transfer learning.

\paragraph{Datasets} ZINC dataset is used for pre-training which consists of 2 million unlabeled molecules sampled from ZINC15~\cite{sterling2015zinc}. Other eight datasets are downstream datasets contained in MoleculeNet~\cite{wu2018moleculenet}, input node features are atom number and chirality tag, and edge features are bond type and direction. Scaffold-split is used to splits graphs into train/val/test sets that mimic real-world use cases. The statistics of these datasets can be found in Table~\ref{tab:mole}. These datasets are widely used for evaluating the transferability of pre-training methods~\cite{you2020graph,infograph,you2021graph}.
\begin{table*}[htp]
\centering
\begin{tabular}{c|ccccccccc}
\hline
\hline
                           & ZINC      & BBBP  & Tox21 & ToxCast & SIDER & ClinTox & MUV    & HIV    & BACE  \\ \hline\hline
\# graphs                  & 2,000,000 & 2,039 & 7,831 & 8,576   & 1,427 & 1,477   & 93,087 & 41,127 & 1,513 \\
\# binary prediction tasks & -         & 1     & 12    & 617     & 27    & 2       & 17     & 1      & 1     \\
Avg. \# nodes              & 26.6      & 24.1  & 18.6  & 18.8    & 33.6  & 26.2    & 24.2   & 24.5   & 34.1  \\ \hline
\end{tabular}
\caption{Data statistics of datasets used for molecular property prediction. ZINC dataset is used for pre-training.}
\label{tab:mole}
\end{table*}

\begin{table*}[htp]
\centering
\begin{tabular}{l|cccccccc|c}
\hline
\hline
                                    & BBBP     & Tox21    & ToxCast  & SIDER    & ClinTox  & MUV      & HIV      & BACE     & Avg. \\ 
\hline
\hline
No-pretrain  & 65.5±1.8 & 74.3±0.5 & 63.3±1.5 & 57.2±0.7 & 58.2±2.8 & 71.7±2.3 & 75.4±1.5 & 70.0±2.5 & 67.0 \\ \hline
ContextPred  & 64.3±2.8 & 75.7±0.7 & 63.9±0.6 & 60.9±0.6 & 65.9±3.8 & 75.8±1.7 & 77.3±1.0 & 79.6±1.2 & 70.4 \\
AttrMasking  & 64.3±2.8 & 76.7±0.4 & 64.2±0.5 & 61.0±0.7 & 71.8±4.1 & 74.7±1.4 & 77.2±1.1 & 79.3±1.6 & 71.1 \\
Infomax      & 68.8±0.8 & 75.3±0.5 & 62.7±0.4 & 58.4±0.8 & 69.9±3.0 & 75.3±2.5 & 76.0±0.7 & 75.9±1.6 & 70.3 \\
GraphCL      & 69.7±0.7 & 73.9±0.7 & 62.4±0.6 & 60.5±0.9 & 76.0±2.7 & 69.8±2.7 & 78.5±1.2 & 75.4±1.4 & 70.8 \\
JOAO         & 70.2±1.0 & 75.0±0.3 & 62.9±0.5 & 60.0±0.8 & 81.3±2.5 & 71.7±1.4 & 76.7±1.2 & 77.3±0.5 & 71.9 \\
GraphLoG     & 72.5±0.8 & 75.7±0.5 & 63.5±0.7 & 61.2±1.1 & 76.7±3.3 & 76.0±1.1 & 77.8±0.8 & 83.5±1.2 & 73.4 \\ 
GraphMAE     & 72.0±0.6 & 75.5±0.6 & 64.1±0.3 & 60.3±1.1 & 82.3±1.2 & 76.3±2.4 & 77.2±1.0 & 83.1±0.9 & 73.8 \\ \hline
\ours(Ours)       & \textbf{72.6±0.4} & \textbf{76.7±0.4} & \textbf{64.3±0.2} & \textbf{62.6±0.4} & \textbf{83.3±0.9} & \textbf{79.8±1.3} & \textbf{78.7±0.4} & \textbf{84.3±0.4} & \textbf{75.3}  \\
\hline
\end{tabular}
\caption{Results on downstream molecular property prediction benchmarks. Avg represents the average performance of each method.}
\label{tab:transfer}
\end{table*}

\paragraph{Baselines} We employ several baselines containing no pre-trained GIN (\ie, directly fine-tune on downstream dataset without self-supervised pre-training), as well as GraphCL~\cite{you2020graph}, JOAO~\cite{you2021graph}, GraphLoG~\cite{graphlog}, GraphMAE~\cite{gmae} and three different pre-training strategies (\ie, Infomax, AttrMasking and ContextPred) proposed in \cite{hu2020strategies} which incorporates the domain knowledge heuristically that correlates with the specific downstream datasets. All baselines (except No-pretrain) will firstly be pre-trained on ZINC datasets and then adopt them on downstream datasets.

\paragraph{Experimental Setup} We evaluate \ours under the transfer learning setting as follows~\cite{gmae,you2020graph}. Firstly, we pre-train the model with 2 million unlabeled molecules sampled from the ZINC15\cite{sterling2015zinc}, and then we finetune the pre-trained models in 8 multi-label multi-class benchmark datasets contained in MoleculeNet~\cite{wu2018moleculenet} with scaffold-split. In our experiments, in the local branch, we will reconstruct node features and not reconstruct edge features like \cite{gmae}. In the global branch, we set dynamic queue size $Q$, temperature and momentum as 4096, 0.05 and 0.999 separately. And for simplicity, we do not use additional augmentation here. For evaluation, we run experiments 10 times and report the mean and standard deviation of ROC-AUC scores(\%) on 8 downstream datasets following~\cite{gmae,you2020graph}.

\paragraph{Analysis} From Table~\ref{tab:transfer}, you can find \ours surpasses other methods and reaches SOTA performance on eight datasets. It reaches around 8\% improvement on No-pretrain GIN which proves our pre-training method \ours is strong enough. And it can enhance GraphMAE a lot on many datasets (\eg, 3.5\% and 2.3\% gains on MUV and SIDER datasets) which further shows the effectiveness of combining contrastive and generative methods. 

\subsection{Node Classification}
\begin{table}[htb]
\centering
\begin{tabular}{cccccc}
\toprule
Dataset     & \# N & \# E & \# F & \# C & H \\ 
\midrule
Cora        & 2078     & 5278     & 1433        & 7          & 0.81        \\
CiteSeer    & 3327     & 4676     & 3703        & 6          & 0.74        \\
PubMed      & 19717    & 44327    & 500         & 3          & 0.80        \\
\bottomrule
\end{tabular}
\caption{Details of used datasets, where we substitute N for \emph{Nodes}, E for \emph{Edges}, F for \emph{Features}, C for \emph{Classes}, H for \emph{Homophily ratio}.}
\label{app:citation}
\end{table}
\paragraph{Datasets} We choose three commonly used citation datasets, \ie, Cora, CiteSeer and PubMed~\cite{sen2008collective}, in this part. These datasets consist of nodes representing different papers and edges signifying citation relationships between them. Each node is associated with a bag-of-words representation of the corresponding paper, and the label corresponds to the academic topic of the paper. Further details about the dataset statistics can be found in Table~\ref{app:citation}.
\paragraph{Experimental Setup} We follow the same experimental setup in GraphMAE~\cite{gmae}. After finishing unsupervised pre-training, we will freeze the pre-trained model and exclusively train a linear classifier using labeled training data for downstream tasks. The specific hyperparameters are listed in Table~\ref{tab:hyper_node}. And we report the mean accuracy with standard deviation after 20 runs. For simplicity, we only compare with GraphMAE here because it is a strong baseline.

Furthermore, in this section, we assess the performance of our prompt method in a few-shot setting\footnote{In order to achieve prompt tuning on node classification, we will extract ego graphs for each target node and subsequently assign ego graph labels based on the corresponding node labels. The subsequent steps mirror the graph-level process.}. The model configuration remains consistent with the above. For standard fine-tuning approaches, all parameters are fine-tuned using downstream data. In contrast, for prompt methods, pre-trained models are kept frozen, with only additional parameters for the downstream task will be tuned.

\begin{table}[htp]
\scalebox{1}{
\begin{tabular}{c|c|ccc}
\hline
\hline
                                     &                & Cora & CiteSeer & PubMed  \\
\hline
\hline
\multirow{6}{*}{\makecell[c]{Model}} 
                                     & type           & GAT      & GAT   & GAT   \\
                                     & hidden\_size   & 512      & 512   & 1024   \\
                                     & num\_head      & 4        & 2     & 4     \\
                                     & num\_layer     & 2        & 2     & 2     \\
                                     & activation     & prelu    & prelu & prelu  \\
                                     & norm           & BN       & BN    & BN    \\
\hline
\multirow{3}{*}{Local}        & scaling factor        & 3        & 3     & 3     \\
                                     & masking rate   & 0.5      & 0.5   & 0.75   \\
                                     & replace rate   & 0.05     & 0.1   & 0.0    \\
\hline
\multirow{8}{*}{Global}              & $Q$            & 1024     & 102400& 1024 \\
                                     & momentum       & 0.999    & 0.999 & 0.999    \\
                                     & tempurate      & 2        & 2     & 2       \\
                                     & feat\_mask1    & 0.0      & 0.1   & 0.3   \\
                                     & feat\_mask2    & 0.1      & 0.5   & 0.3   \\
                                     & drop\_edge1    & 0.1      & 0.0   & 0.2   \\
                                     & drop\_edge2    & 0.5      & 0.1   & 0.2   \\
                                     & loss coeff.    & 0.1      & 0.08  & 0.1   \\
\hline
\multirow{5}{*}{Training}            
                                     & epochs         & 1500     & 300   & 1000       \\
                                     & learning rate  & 0.001    & 0.001 & 0.001     \\
                                     & weight\_decay  & 2e-4     & 2e-5  & 1e-5   \\
                                     & optimizer      & Adam     & Adam  & Adam     \\
                                     & scheduler      & True     & True  & True    \\
\hline
\end{tabular}}
\caption{Hyper-parameters for pre-training method.}
\label{tab:hyper_node}
\end{table}

\paragraph{Analysis} Table~\ref{tab:node} shows the results of unsupervised representation learning and illustrates the consistent superiority of our \ours pre-training method across these three datasets, which proves the effectiveness of combining contrastive and generative methods. 

Moreover, our prompt tuning approach for node classification also demonstrates its effectiveness, as seen in Table~\ref{tab:node_prompt}. Notably, our prompt method achieves over 3\% absolute improvement compared to standard fine-tuning approaches and over 2.4 \% absolute improvement compared to another graph prompt method, GPPT, on 5-shot Cora dataset. These results show the superiority of our prompt design even on node-level tasks.
\begin{table}[htp]
\centering
\scalebox{1}{
\begin{tabular}{lccc}
\toprule
         & Cora       & CiteSeer    & PubMed      \\
\midrule
GraphMAE & 83.77±0.62 & 73.04±0.28  & 81.05±0.30  \\
SGL      & \textbf{84.15±0.35} & \textbf{73.48±0.27}  & \textbf{81.31±0.41}\\
\bottomrule
\end{tabular}}
\caption{The mean accuracy with standard deviation of unsupervised representation learning for node classification with 20 different random seeds.}
\label{tab:node}
\end{table}

\begin{table}[htp]
\centering
\scalebox{0.85}{
\begin{tabular}{l|cc|cc}
\toprule
         & \multicolumn{2}{c|}{Cora}& \multicolumn{2}{c}{PubMed}      \\
         &3-shot       & 5-shot     & 3-shot     & 5-shot       \\
\midrule
GraphMAE & 68.50±1.67  & 72.48±1.08 & 68.80±0.36 & 73.08±0.56  \\
SGL      & 70.32±0.74  & 74.14±0.82 & 68.98±0.62 & 73.72±0.77  \\ \midrule
GPPT     & 72.14±2.64  & 74.80±1.76 & 69.52±0.80 & 72.50±0.93   \\
SGL-PT   & \textbf{73.84±1.89}  & \textbf{77.26±0.98} & \textbf{70.08±1.71} & \textbf{75.50±0.90}   \\
\bottomrule
\end{tabular}}
\caption{The mean accuracy with standard deviation of FT \& PT for node classification. GPPT uses SGL as its pre-trained model in this table.}
\label{tab:node_prompt}
\end{table}


\subsection{Evaluating \ourpts on Other Graph-level Datasets}
Due to space constraints, only partial results of prompt tuning are presented in the main content. In this section, additional experiments are exhibited in Table~\ref{tab:other_graph_prompt}. The conclusions align with those stated in the main paper.

\begin{table}[htp]
\centering
\scalebox{0.85}{
\begin{tabular}{l|cc|cc}
\toprule
         & \multicolumn{2}{c|}{MOLT-4}& \multicolumn{2}{c}{P388}    \\
         & 1-shot     & 3-shot      & 1-shot & 3-shot               \\
\midrule
GraphMAE & 52.80±1.85 & 52.90±1.64  & 52.45±2.92 & 53.39±1.41             \\
SGL      & 52.67±2.31 & 52.75±2.80  & 53.96±3.37 & 54.66±1.53                  \\ \midrule
GPPT     & 50.08±0.68 & 50.77±1.23  & 50.37±2.40 & 52.52±3.77                      \\
SGL-PT   & \textbf{52.97±2.10} & \textbf{54.68±1.72}  & \textbf{54.89±1.87} & \textbf{55.33±2.13}               \\
\bottomrule
\end{tabular}}
\caption{The mean ROC-AUC with standard deviation of FT \& PT for graph classification on other datasets. GPPT uses SGL as its pre-trained model in this table.}
\label{tab:other_graph_prompt}
\end{table}

\subsection{Applying \ourpts on Other Pre-trained Methods}
Our prompt method can also be applied to other contrastive methods, because our prompt loss is $\mathcal{L}_{\text{prompt}} = \lambda_{\text{prompt}}\cdot\mathcal{L}_{\text{local}} + \left(1-\lambda_{\text{prompt}}\right)\cdot\mathcal{L}_{\text{proto}}$, by setting the coefficient $\lambda_{\text{prompt}}$ as 0, our prompt aligns the objectives of pre-text and downstream tasks (\ie, contrastive loss). To validate this assertion, we substitute the pre-trained model with the GraphCL model. The outcomes are detailed in Table~\ref{tab:other_model}. Our approach consistently outperforms standard fine-tuning by a significant margin. For instance, on 3-shot DD dataset, \ourpts surpasses GraphCL by more than 6\% absolute improvement, and on 3-shot MUTAG dataset, it shows over 12\% absolute improvement. Furthermore, our prompt method achieves superior results compared to other graph prompt methods, underscoring the efficacy of our prompt design. This success also validates the adaptability of our prompt approach to other contrastive methods.

\begin{table}[htp]
\centering
\scalebox{0.85}{
\begin{tabular}{l|cc|cc}
\toprule
         & \multicolumn{2}{c|}{DD}    & \multicolumn{2}{c}{MUTAG}      \\
                             & 1-shot     & 3-shot & 1-shot & 3-shot  \\
\midrule
$\text{GraphCL}^{\diamond}$  & 52.30±2.51 & 53.39±2.87 & 57.92±3.43 & 62.39±1.53   \\\midrule
$\text{GPPT}^{\diamond}$     & 54.11±2.09 & 58.31±2.40 & 65.46±5.02 & 69.51±2.56  \\
$\text{GPF}^{\diamond}$      & 55.66±2.20 & 58.30±1.26 & 59.43±3.40 & 64.07±2.38  \\
$\text{SGL-PT}^{\diamond}$   & \textbf{57.15±1.53} & \textbf{59.49±1.25} & \textbf{68.58±5.49} & \textbf{74.47±2.32}  \\
\bottomrule
\end{tabular}}
\caption{The mean accuracy with standard deviation of FT \& PT for graph classification. $\diamond$ means we use GraphCL as their pre-trained models.}
\label{tab:other_model}
\end{table}

\subsection{Sensitivity Analysis}
\begin{figure}
    \centering
    \includegraphics[scale=0.28]{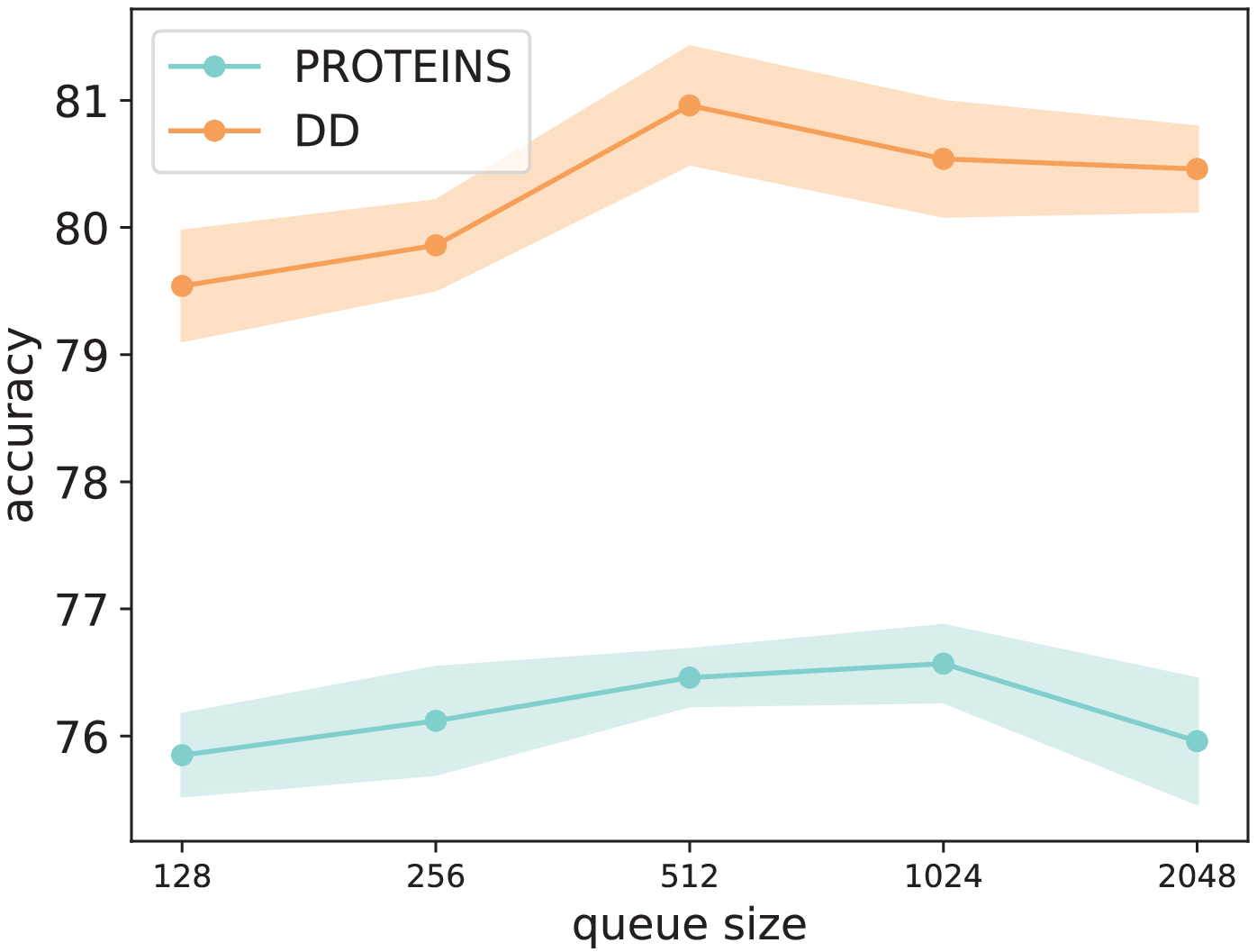}
    \includegraphics[scale=0.28]{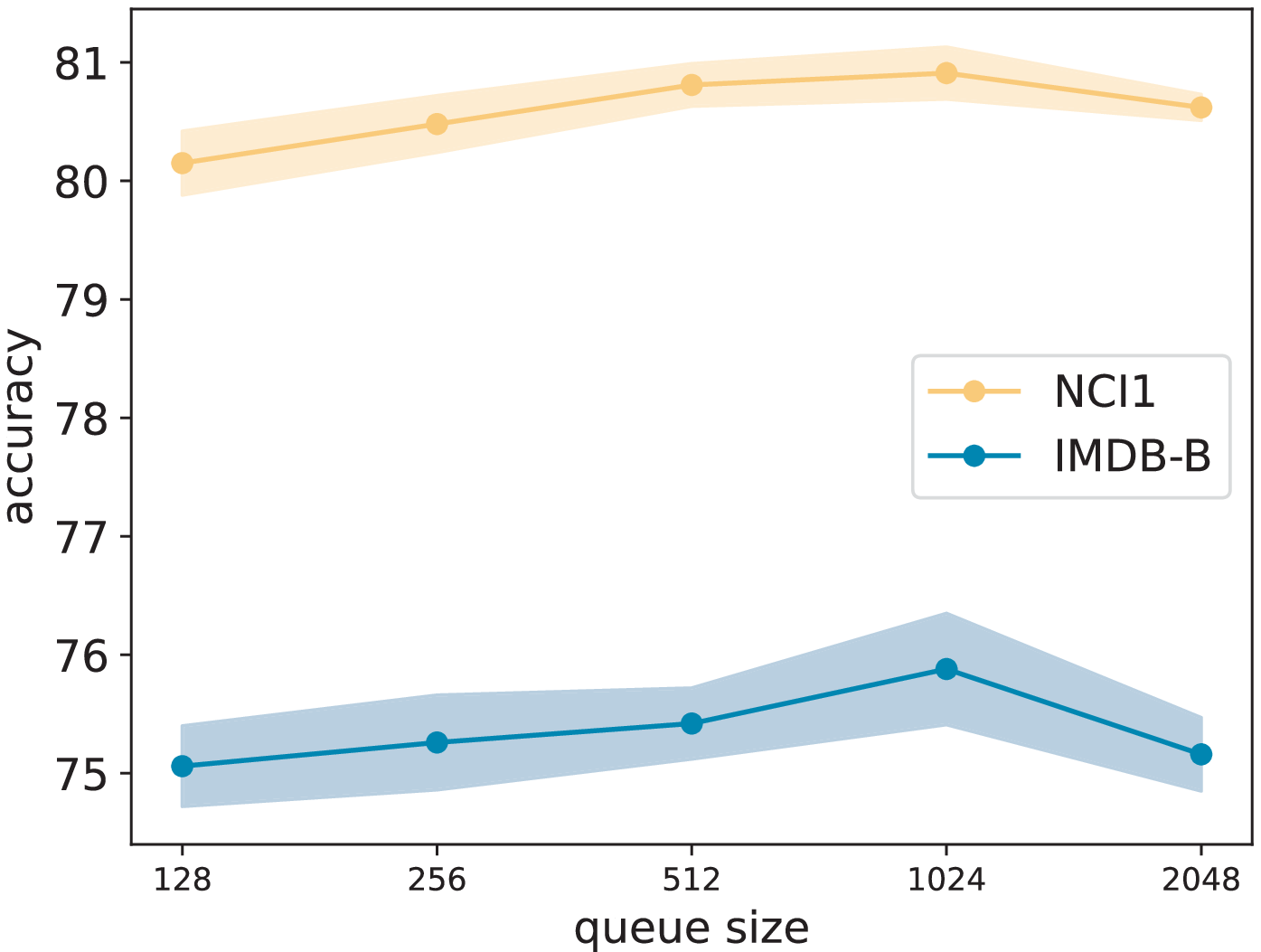}
    \caption{Sensitivity analysis of the dynamic queue $Q$'s size.}
    \label{fig:sentivity}
\end{figure}

\paragraph{Analysis on $Q$} In the ablation study, we find that the dynamic queue is essential to integrate contrastive and generative methods better. 
In this section, we dig into the impact of the size $Q$ of the dynamic queue. The results are shown in Figure \ref{fig:sentivity}, and we can find that the performance moves up with the increase of $Q$ and the gets best performance when $Q$ reaches 512 or 1024. This size is much larger than batch size (\eg, 8, 16), which shows the necessity of the dynamic queue to integrate the two branches. 


\paragraph{Analysis on $\lambda$}
During pre-training, the model captures both local and global information, so we set $\lambda_\text{pre}$ as 0.5. During prompt tuning, we focus on learning class prototypes used for classification, so we lower the ratio of $\mathcal{L}_\text{local}$ ($\lambda_\text{prompt}$ as 0.1). Empirically, we found the performance of downstream tasks will drop with $\lambda_\text{prompt}$ increasing (for DD, $\lambda_\text{prompt}$=$0.0,\mathbf{0.1},0.3,0.5,0.7,0.9$. Acc=$74.3_{0.5},\mathbf{75.4_{0.2}},74.5_{0.5},73.3_{0.5},73.2_{0.4},73.0_{0.4}$). So, we set $\lambda_\text{prompt}$ as 0.1 for all datasets.

\begin{table}[htp]
    \centering
    \begin{tabular}{l|cccc}
    \toprule
\multirow{2}{*}{Method}& \multicolumn{1}{c}{DD}   &\multicolumn{1}{c}{MUTAG}& \multicolumn{1}{c}{COLLAB}            \\
               & Params(K)  & Params(K) & Params(K)  \\ \midrule
        GIN    & 820        & 10        & 296        \\ \midrule
        GPPT   & 258        & 1         & 64          \\
        GPF    & 1.10       & 0.07      & 1.10         \\
        ProG   & 1.87       & 0.13      & 4.67         \\
   GraphPrompt & 0.50       & 0.03      & 0.25         \\
        SGL-PT & 1          & 0.06      & 0.75         \\ \bottomrule
    \end{tabular}
    \caption{Study of parameter efficiency.}
    \label{tab:params}
\end{table}

\section*{D. Parameter Efficiency} We compare trainable parameters with GIN, GPPT, GPF, ProG and GraphPrompt and \ourpts in Table~\ref{tab:params}. The `GIN' entry represents the number of parameters required to train a GIN model from scratch, which is significantly more than other methods. For prompt methods, we will freeze the pre-trained models and only calculate the downstream task-specific parameters. GPPT requires task tokens (used for classification) and a structure token involving attention module (used for neighbor aggregation) to accomplish downstream tasks, resulting in significantly more trainable parameters than other prompt methods. For GPF, it needs to initialize a learnable vector as well as a downstream classifier. For ProG, it needs to initialize a learnable prompt graph (\ie, a set of learnable vectors). For GraphPrompt, it introduces a learnable weighted summation readout function but needs to execute clustering to obtain class prototypes. As for \ourpt, we only require training class prototypes for classification. 

GraphPrompt, while having the fewest trainable parameters, does require additional computational time due to its clustering process for obtaining class prototypes. This step becomes increasingly time-consuming as the downstream data expands. For example, in semi-supervised settings, executing GraphPrompt on NCI-H23 takes more than a day, while other methods complete the task in just two hours.

To conclude, while GraphPrompt has fewer trainable parameters than \ourpt, the efficiency and effectiveness of \ourpts surpass GraphPrompt.